# Event-Based Dense Reconstruction Pipeline

Kun Xiao[1], Guohui Wang[2], Yi Chen[1], Jinghong Nan[1], Yongfeng Xie[1]
[1] Beijing Institute of Aerospace Systems Engineering, Beijing, China
[2] China Academy of Launch Vehicle Technology, Beijing, China
e-mail: robin_shaun@foxmail.com

*Abstract*—Event cameras are a new type of sensors that are different from traditional cameras. Each pixel is triggered asynchronously by event. The trigger event is the change of the brightness irradiated on the pixel. If the increment or decrement of brightness is higher than a certain threshold, an event is output. Compared with traditional cameras, event cameras have the advantages of high dynamic range and no motion blur. Since events are caused by the apparent motion of intensity edges, the majority of 3D reconstructed maps consist only of scene edges, i.e., semi-dense maps, which is not enough for some applications. In this paper, we propose a pipeline to realize event-based dense reconstruction. First, deep learning is used to reconstruct intensity images from events. And then, structure from motion (SfM) is used to estimate camera intrinsic, extrinsic and sparse point cloud. Finally, multi-view stereo (MVS) is used to complete dense reconstruction.

*Keywords—Event camera, dense reconstruction, deep learning, structure from motion, multi-view stereo*

## I. Introduction

Recent advances in 3D sensors have, more than ever, increased our access to 3D information, empowering high quality 3D reconstruction, scene/model recreation and human-machine interaction. Although visual 3D reconstruction develops rapidly, high dynamic range (HDR) scene and high speed motion still pose challenges.

Event cameras are a new type of bionic sensor that has become popular in academia in the past decade. Compared with traditional cameras, the way it acquires visual information has changed. The event camera does not trigger all pixels synchronously, but each pixel is triggered asynchronously by events [1, 2]. The trigger event is the change of the brightness of the light shining on the pixel, and if the increment or decrement is higher than a certain threshold, an event is output. Their advantages are: ultra-high temporal resolution (microseconds) and extremely low latency (sub-milliseconds), high dynamic range (120 dB vs. 60 dB for conventional cameras) and low power consumption. Therefore, event cameras have great potential in HDR and high speed motion applications.

Nowadays, 3D reconstruction with event cameras attracts the attention of researchers. Depth estimation with a single event camera has been shown in [3], [4], [5]. These methods recover a semi-dense 3D reconstruction of the scene (i.e., 3D edge map) by integrating information from the events of a moving camera over time. There is no absolute scale in 3D reconstruction with one camera, so a stereo depth estimation method for SLAM has been presented [6]. It obtains a semi-dense 3D reconstruction of the scene by optimizing the local spatio-temporal consistency of events across image planes using time surfaces.

Since events are caused by the apparent motion of intensity edges, the majority of 3D reconstructed maps consist only of scene edges, i.e., semi-dense maps, which is not enough for some applications. At present, deep learning is the only reliable way to realize event-based dense reconstruction. A recurrent neural network architecture is proposed to generate dense depth predictions using a monocular setup [7]. E3D [8] first introduces an event-to-silhouette (E2S) neural network module to transform a stack of event frames to the corresponding silhouettes, with additional neural branches for camera pose regression, and then employs a 3D differentiable renderer (PyTorch3D) to enforce cross-view 3D mesh consistency and fine-tune the E2S and pose network.

In this work, we introduce a new event-based dense reconstruction pipeline, shown as Fig. 1. First, deep learning is used to reconstruct intensity images from events. And then, structure from motion (SfM) is used to estimate camera intrinsic, extrinsic and sparse point cloud. Finally, multi-view stereo (MVS) is used to complete dense reconstruction. Experiment comparisons between the proposed method and SfM + MVS using traditional cameras are conducted, proving the proposed method works better at high dynamic range (HDR) scene or with high speed motion.

## II. Event Representation and Intensity Image Reconstruction

Unlike traditional cameras, which acquire the entire image at a specific sampling frequency (such as 30Hz) through an external clock, event cameras output signals asynchronously and independently according to changes of brightness ((log intensity). The data flow rate of this signal is continuously changing, called "events", and each event represents that the brightness change felt by a certain pixel at a certain moment reaches a threshold, as shown in Fig. 2. This encoding is inspired by the pulsatile nature of biological visual pathways.

When a pixel emits an event, the current brightness is memorized and changes in brightness are continuously monitored. When the change value exceeds a threshold, the camera generates an event, the event contains the pixel coordinates $x$, $y$, time $t$ and the changed polarity $p$ (+1 for brightness increase, -1 for brightness decrease).

Let the brightness $L(\mathbf{x}, t) = \log I(\mathbf{x}, t)$ be the log intensity of the pixel, when the brightness change of a pixel $\mathbf{x}_k = (x_k, y_k)$ of the event reaches a threshold C, the event camera output event $\mathbf{e}_k = (\mathbf{x}_k, t_k, p_k)$, as shown in the formula:

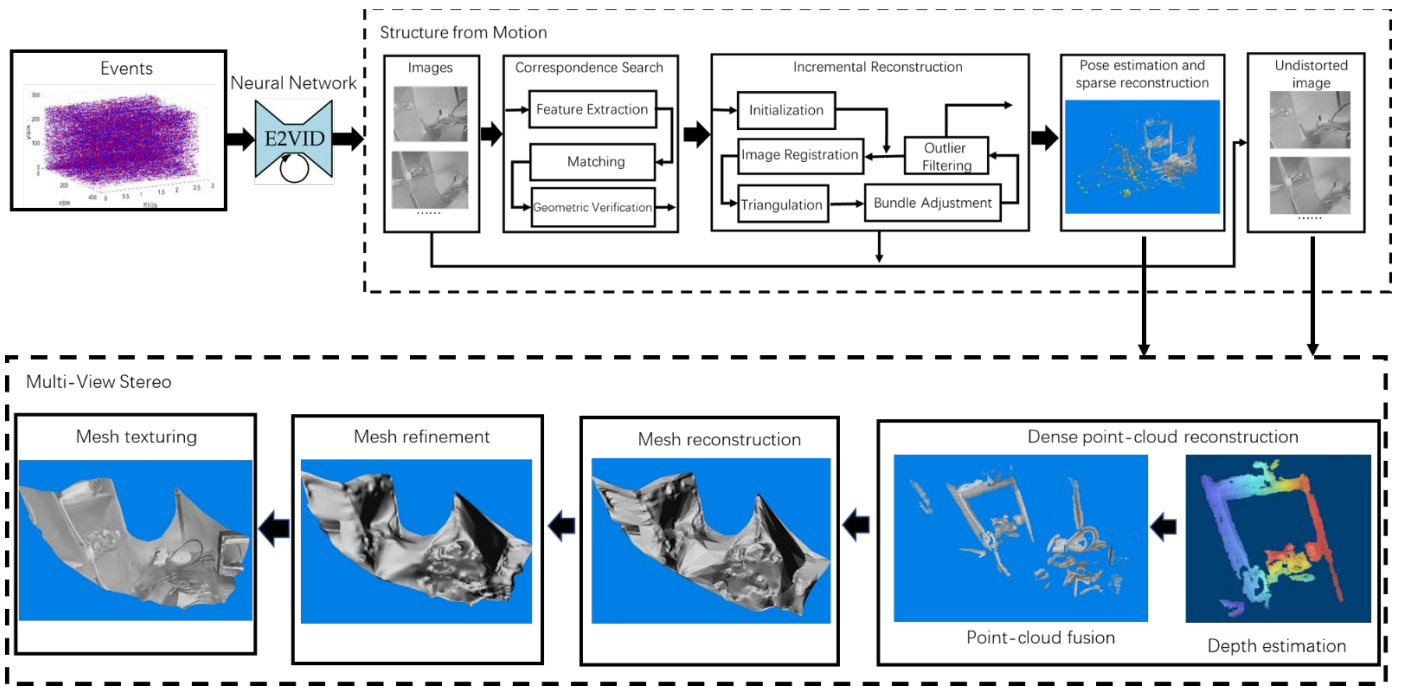

Fig. 1. Event-Based Dense Reconstruction Pipeline

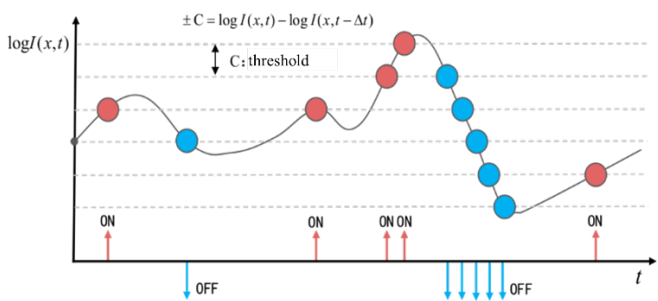

Fig. 2. Schematic diagram of event generation from a single pixel

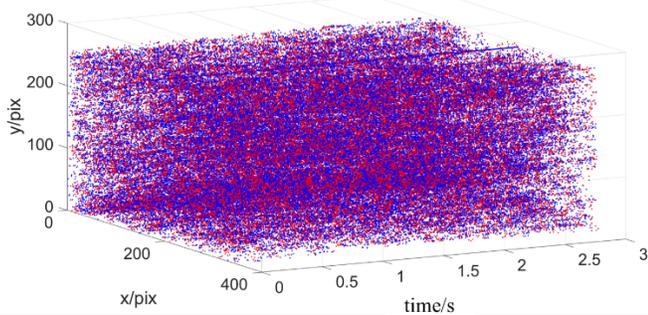

Fig. 3. Events in space time, colored according to polarity (positive in blue, negative in red)

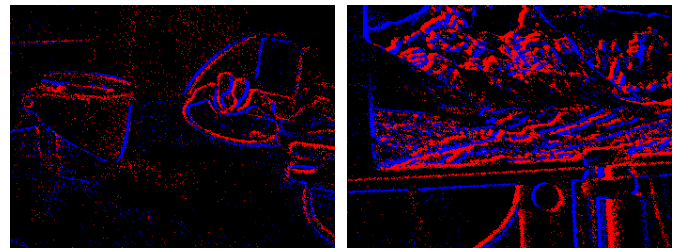

Fig. 4. Event frames under fast motion (left) and HDR (right). Red pixels are positive events and blue pixels are negative events.

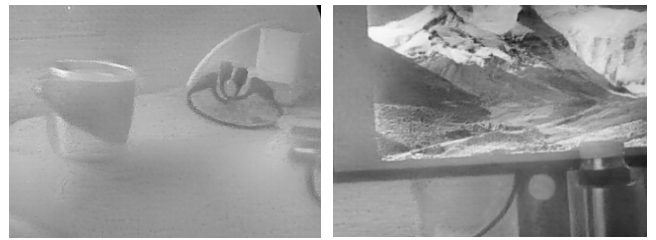

Fig. 5. Reconstructed intensity images under fast motion (left) and high dynamic range scene (right)

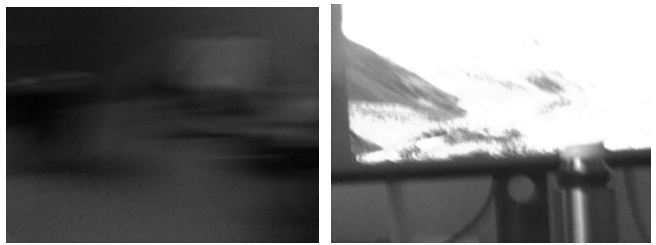

Fig. 6. Images from a traditional camera under fast motion (left) and high dynamic range scene (right)

$$\Delta L(\mathbf{x}_k, t_k) = L(\mathbf{x}_k, t_k) - L(\mathbf{x}_k, t_k - \Delta t_k) = p_k C$$

where $\Delta t_k$ is the duration from the previous event at the $\mathbf{x}_k$ position to the current event.

Events flow continuously along the time axis, forming the spatiotemporal distribution of events, as shown in Fig. 3. This form is very different from traditional images, in order to adapt to the current 3D reconstruction architecture, the event stream needs to be processed. The most direct processing method is to superimpose all events in a time window on a two-dimensional image to form an event frame, as shown in Fig. 4. This processing unit is called an event accumulator [9]. However, because events are usually generated from moving edges, the event frame only presents edges of the object, so that a large amount of scene information is lost. Although theoretically, more scene information can be obtained by means of long-time integration to recover the original intensity image, the actual event generation is also related to the upper limit of transmission bandwidth [10] and temperature [11], so the effect of direct integration is limited.

The problem of obtaining the intensity image from event stream is called the image reconstruction problem. The current mainstream solution is deep learning [12], [13], [14]. In this research, E2VID proposed in [12] is chosen in the proposed pipeline for its better performance and good open-source community. The reconstructed intensity images under fast motion and high dynamic range scene are shown in Fig. 5. Compared to event frames, all scene information is recovered and can be used for 3D dense reconstruction. And compared to images from a traditional camera, shown as Fig. 6, the reconstructed images have better quality.

The input of E2VID is a continuous stream of events, and the output is a normalized image sequence $\{\hat{\mathbf{I}}_k\}$, where $\hat{\mathbf{I}}_k \in [0,1]^{w \times h}$, where $w$ and $h$ are the width and height of the image, respectively. To achieve this, the incoming event stream is divided into event windows. E2VID supports creating event windows by number of events, and creating by duration. Different from the event window required by the SLAM system, E2VID has a high requirement for the continuity of events, so the method of creating event windows by time and number [9] cannot be used. E2VID works better with windows created by the number of events, so we adopt this method to create $\mathbf{W}_k = \{\mathbf{e}_i\}$, where $i \in [0, N-1]$, i.e. each event window contains a fixed number of $N$ events. The reconstruction function is implemented by a recurrent convolutional neural network, as shown in Fig. 7. For each new event window $\mathbf{W}_k$, E2VID generates a new image $\hat{\mathbf{I}}_k$ with the previous state $\mathbf{s}_{k-1}$, and updates the current state $\mathbf{s}_k$. The network is trained using supervised learning, and the dataset is derived from simulator-generated event sequences and corresponding images [15].

Recurrent convolutional networks require a fixed input tensor size, so events in the event window need to be processed. We encode event as a spatio-temporal voxel grid [16]. The time length $\Delta T = t_{N-1}^k - t_0^k$ of the window $\mathbf{W}_k$ is discretized into B temporal bins. Each event distributes its polarity $p_i$ to the two closest spatio-temporal voxels as follows:

$$\mathbf{E}(x_l, y_m, t_n) = \sum_{\substack{x_i = x_l \\ y_i = y_m}} p_i \max(0, 1 - |t_n - t_i^*|)$$

where $t_i^* = \frac{B-1}{\Delta T}(t_i - t_0)$ is the normalized event timestamp, and B=5 is used in this research.

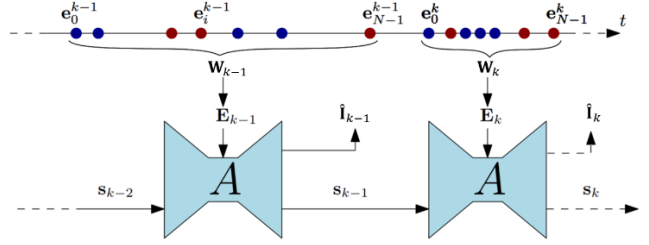

Fig. 7. Overview of E2VID. The event stream (depicted as red/blue dots on the time axis) is split into windows $\varepsilon_k$ containing multiple events. Each window is converted into a 3D event tensor $\mathbf{E}_k$ and passed through the network, together with the previous state $\mathbf{s}_{k-1}$ to generate a new image reconstruction $\hat{\mathbf{I}}_k$ and updated state $\mathbf{s}_k$. In this example, each window $\mathbf{W}_k$ contains a fixed number of events $N = 7$.

## III. STRUCTURE FROM MOTION

The output images of E2VID are asynchronous, but this does not matter for SfM, because SfM is an offline algorithm and does not care timestamps of images.

SfM is the process of reconstructing 3D structure from its projections into a series of images taken from different viewpoints. Incremental SfM (denoted as SfM in this paper) is a sequential processing pipeline with an iterative reconstruction component (Fig. 1) [17], which is chosen in the proposed pipeline. It commonly starts with feature extraction and matching, followed by geometric verification. The resulting scene graph serves as the foundation for the reconstruction stage, which seeds the model with a carefully selected two-view reconstruction, before incrementally registering new images, triangulating scene points, filtering outliers, and refining the reconstruction using bundle adjustment (BA).

### A. Correspondence Search

The first stage is correspondence search which finds scene overlap in the input images $\mathcal{I} = \{I_i \mid i = 1 \dots N_I\}$ and identifies projections of the same points in overlapping images. The output is a set of geometrically verified image pairs $\bar{\mathcal{C}}$ and a graph of image projections for each point.

**Feature Extraction.** For each image $I_i$, SfM detects sets $\mathcal{F}_i = \{(\mathbf{x}_j, \mathbf{f}_j) \mid j = 1 \dots N_{F_i}\}$ of local features at location $\mathbf{x}_j \in \mathbb{R}^2$ represented by an appearance descriptor $\mathbf{f}_j$. The features should be invariant under radiometric and geometric changes so that SfM can uniquely recognize them in multiple images [18]. SIFT [19], its derivatives [20], and more recently learned features [21] are the gold standard in terms of robustness. Alternatively, binary features provide better efficiency at the cost of reduced robustness [22]. In the proposed pipeline, SIFT is chosen for its overall performance.

**Matching.** Next, SfM discovers images that see the same scene part by leveraging the features $\mathcal{F}_i$ as an appearance description of the images. The naïve approach tests every image pair for scene overlap; it searches for feature correspondences by finding the most similar feature in image $I_a$ for every feature in image $I_b$, using a similarity metric comparing the appearance $\mathbf{f}_j$ of the features. This approach has computational complexity $O(N_I^2 N_{F_i}^2)$. Because the reconstructed image quality cannot be compared with images from traditional camera under good light condition and static scene (generally settings for high quality 3D reconstruction), large computational complexity is accepted for good matching. The output is a set of potentially overlapping image pairs $C = \{\{I_a, I_b\} \mid I_a, I_b \in \mathcal{I}, a < b\}$ and their associated feature correspondences $\mathcal{M}_{ab} \in \mathcal{F}_a \times \mathcal{F}_b$.

**Geometric Verification.** The third stage verifies the potentially overlapping image pairs $C$. Since matching is based solely on appearance, it is not guaranteed that corresponding features actually map to the same scene point. Therefore, SfM verifies the matches by trying to estimate a transformation that maps feature points between images using projective geometry. Depending on the spatial configuration of an image pair, different mappings describe their geometric relation. A homography $\mathbf{H}$ describes the transformation of a purely rotating or a moving camera capturing a planar scene [23]. Epipolar geometry [23] describes the relation for a moving camera through the essential matrix $\mathbf{E}$ (calibrated) or the fundamental matrix $\mathbf{F}$ (uncalibrated), and can be extended to three views using the trifocal tensor [23]. If a valid transformation maps a sufficient number of features between the images, they are considered geometrically verified. Since the correspondences from matching are often outlier-contaminated, robust estimation techniques, such as RANSAC [24] in the proposed pipeline, are required. The output of this stage is a set of geometrically verified image pairs $\bar{C}$, their associated inlier correspondences $\bar{\mathcal{M}}_{ab}$, and optionally a description of their geometric relation $G_{ab}$. The output of this stage is a so-called scene graph [25] with images as nodes and verified pairs of images as edges.

*B. Incremental Reconstruction*

The input to the reconstruction stage is the scene graph. The outputs are pose estimates $\mathcal{P} = \{\mathbf{P}_c \in \mathbf{SE}(3) \mid c = 1 \ldots N_P\}$ for registered images and the reconstructed scene structure as a set of points $\mathcal{X} = \{\mathbf{X}_k \in \mathbb{R}^3 \mid k = 1 \ldots N_X\}$.

**Initialization.** SfM initializes the model with a carefully selected two-view reconstruction [26]. Choosing a suitable initial pair is critical, since the reconstruction may never recover from a bad initialization. Moreover, the robustness, accuracy, and performance of the reconstruction depends on the seed location of the incremental process. Initializing from a dense location in the image graph with many overlapping cameras typically results in a more robust and accurate reconstruction due to increased redundancy. In contrast, initializing from a sparser location result in lower runtimes, since BAs deal with overall sparser problems accumulated over the reconstruction process.

**Image Registration.** Starting from a metric reconstruction, new images can be registered to the current model by solving the Perspective-n-Point (PnP) problem [24] using feature correspondences to triangulated points in already registered images (2D-3D correspondences). The PnP problem involves estimating the pose $\mathbf{P}_c$ and, for uncalibrated cameras, its intrinsic parameters. The set $\mathcal{P}$ is thus extended by the pose $\mathbf{P}_c$ of the newly registered image. A novel robust next best image selection method from [17] for accurate pose estimation and reliable triangulation is used in the proposed pipeline.

**Triangulation.** A newly registered image must observe existing scene points. In addition, it may also increase scene coverage by extending the set of points $\mathcal{X}$ through triangulation. A new scene point $\mathbf{X}_k$ can be triangulated and added to $\mathcal{X}$ as soon as at least one more image, also covering the new scene part but from a different viewpoint, is registered. Triangulation is a crucial step in SfM, as it increases the stability of the existing model through redundancy [27] and it enables registration of new images by providing additional 2D-3D correspondences. A robust and efficient triangulation method from [15] is used in the proposed pipeline.

**Bundle Adjustment.** Image registration and triangulation are separate procedures, even though their products are highly correlated - uncertainties in the camera pose propagate to triangulated points and vice versa, and additional triangulations may improve the initial camera pose through increased redundancy. Without further refinement, SfM usually drifts quickly to a non-recoverable state. BA [27] is the joint non-linear refinement of camera parameters $\mathbf{P}_c$ and point parameters $\mathbf{X}_k$ that minimizes the reprojection error

$$E = \sum_j \rho_j \left( \|\pi(\mathbf{P}_c, \mathbf{X}_k) - \mathbf{x}_j\|_2^2 \right)$$

using a function $\pi$ that projects scene points to image space and a loss function $\rho_j$ to potentially down-weight outliers. Levenberg-Marquardt [23, 27] is the method of choice for solving BA problems.

IV. MULTI-VIEW STEREO

Multi-view stereo aims at recovering the full surface of the scene to be reconstructed. The input is a set of camera poses plus the sparse point-cloud and the output is a textured mesh. Four modules are included: dense point-cloud reconstruction for obtaining a complete and accurate as possible point-cloud, mesh reconstruction for estimating a mesh surface that explains the best the input point-cloud, mesh refinement for recovering all fine details and mesh texturing for computing a sharp and accurate texture to color the mesh.

*A. Dense Point-Cloud Reconstruction*

The goal of this module is to provide the functionality of obtaining a complete and accurate point-cloud at reasonable speeds. Since the final goal is to obtain a mesh representation, and since there is a module to refine the mesh, the completeness and speed of estimating the dense point-cloud is more important than the accuracy. In the proposed pipeline, Patch-Match Stereo [28] is used.

## B. Mesh Reconstruction

This module aims at estimating a mesh surface that explains the input point-cloud, and it should be robust to outliers. The input point-cloud could be not so dense, and the algorithm used should also be able to perform well. In the proposed pipeline, algorithm proposed in [29] is used.

## C. Mesh Refinement

Rough meshes obtained by the previous module are in general a good enough starting point for a variational refinement step. Such algorithms are relatively fast and able to recover the true surface even in cases when only a coarse input mesh is provided (as in the case of meshes estimated from a sparse point-cloud, or texture-less scenes). The algorithm employed for solving this task is based on the [30].

## D. Mesh Texturing

In the case of having a perfect mesh reconstruction and ground-truth camera poses, obtaining the texture is relatively a straightforward step. In reality however both the mesh and the camera poses contain slight variations/errors at best, and hence the mesh texturing module should be able to cope with them. An algorithm proposed in [31] is used in our proposed pipeline.

## V. EXPERIEMNT

In order to verify our proposed pipeline, experiment comparisons between the proposed method and SfM + MVS using traditional cameras are conducted. A DAVIS346 event camera which contains 346×260 pixels is used. Two challenging situations, high speed motion and HDR scene, are provided.

Fig. 8 shows the comparison between event-based and traditional dense reconstruction with high speed motion. It can be seen that the completeness and accuracy of the former are better than latter. Fig. 9 shows the comparison at high dynamic range scene. Similarly, the reconstruction quality of event-based method are better. The key reason is that reconstructed intensity images, shown as Fig. 5, have higher quality than images from a traditional camera, shown as Fig. 6, with high speed motion or at HDR scene.

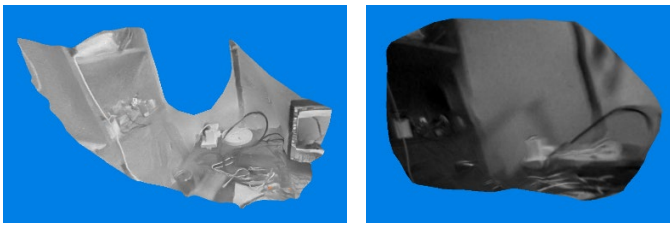

Fig. 8 Event-based (left) and traditional (right) dense reconstruction with high speed motion

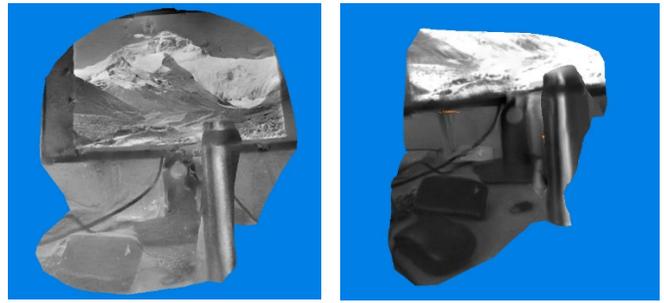

Fig. 9 Event-based (left) and traditional (right) dense reconstruction at high dynamic range scene

## VI. CONCLUSION

In this paper, we propose a pipeline to realize event-based dense reconstruction. First, deep learning is used to reconstruct intensity images from events. And then, structure from motion (SfM) is used to estimate camera intrinsic, extrinsic and sparse point cloud. Finally, multi-view stereo (MVS) is used to complete dense reconstruction. Experiment comparisons between the proposed method and SfM + MVS using traditional cameras are conducted, proving the proposed method works better at high dynamic range (HDR) scene or with high speed motion.

## ACKNOWLEDGMENT

Thanks to COLMAP[1] and OpenMVS[2] for their good open source projects. Thanks to the research team of Professor Xiangke Wang from National University of Defense Technology for providing the DAVIS346 event camera.

---

[1] https://github.com/colmap/colmap
[2] https://github.com/cdcseacave/openMVS